\documentclass{article}

\usepackage{arxiv}

\usepackage[table,xcdraw]{xcolor}
\usepackage{float}

\usepackage[utf8]{inputenc} 
\usepackage[T1]{fontenc}    
\usepackage{hyperref}       
\usepackage{url}            
\usepackage{booktabs}       
\usepackage{amsfonts}       
\usepackage{nicefrac}       
\usepackage{microtype}      
\usepackage{graphicx}
\usepackage{pdfpages} 

\title{Biologically inspired sleep algorithm for artificial neural networks
\thanks{Supported by the Lifelong Learning Machines program from DARPA/MTO (HR0011-18-2-0021)}}

\author{%
 \{Giri P Krishnan$^1$, Timothy Tadros$^1$\}$^{+}$, Ramyaa Ramyaa$^2$, Maxim Bazhenov$^1$ \\
 1. Department of Medicine, University of California, San Diego, CA\\
 2. Department of Computer Science, New Mexico Tech, Socorro, NM\\
 + equal contributions 
}

\begin{document}

\maketitle

\begin{abstract}
  Sleep plays an important role in incremental learning and consolidation of memories in biological systems. Motivated by the processes that are known to be involved in sleep generation in biological networks, we developed an algorithm that implements a sleep-like phase in artificial neural networks (ANNs). After initial training phase, we convert the ANN to a spiking neural network (SNN) and simulate an offline sleep-like phase  using spike-timing dependent plasticity rules to modify synaptic weights. The SNN is then converted back to the ANN and evaluated or trained on new inputs. We demonstrate several performance improvements after applying this processing to ANNs trained on MNIST, CUB200 and a motivating toy dataset. First, in an incremental learning framework, sleep is able to recover older tasks that were otherwise forgotten in the  ANN without sleep phase due to catastrophic forgetting. Second, sleep results in forward transfer learning of unseen tasks. Finally, sleep improves  generalization ability of the ANNs to classify images with various types of noise. We provide a theoretical basis for the beneficial role of the brain-inspired sleep-like phase for the ANNs and present an algorithmic way for future implementations of the various features of sleep in deep learning ANNs. Overall, these results suggest that biological sleep can help mitigate a number of problems ANNs suffer from, such as poor generalization and catastrophic forgetting for incremental learning. 
\end{abstract}

\section{Introduction}
\label{intro}


Although artificial neural networks (ANNs) have equaled and even surpassed human performance on various tasks \cite{russakovsky2015imagenet, silver2018general}, they suffer from a range of problems. First, ANNs suffer from catastrophic forgetting \cite{goodfellow2013empirical, mcclelland1995there}. While humans and animals can continuously learn from new information, ANNs perform well on new tasks while forgetting older tasks that are not explicitly retrained. Second, ANNs fail to generalize to multiple examples of the specific task for which they were trained \cite{geirhos2018generalisation, dodge2017study, szegedy2013intriguing}. Indeed, ANNs are usually trained with highly filtered datasets, which limits the extent to which they can generalize beyond these filtered examples. In contrast, humans robustly act in the presence of limited or altered (e.g., by noise) stimulus conditions \cite{geirhos2018generalisation, dodge2017study}. Thirdly, ANNs sometimes fail to transfer learning to the other similar tasks apart from the ones they were explicitly trained on \cite{pan2009survey}. In contrast, humans represent information in a generalized fashion that does not depend on the exact properties or conditions of how the task was learned  \cite{spengler1997learning}. This allows the mammalian brain to transfer old knowledge to unlearned tasks, while the current state-of-the-art deep learning models are unable to do so. 

Sleep has been hypothesized to play an important role in memory consolidation and generalization of knowledge in biological brain \cite{walker2004sleep,stickgold2013sleep,rasch2013sleep}. 
During sleep, neurons are spontaneously active without external input and generate complex patterns of synchronized oscillatory activity across brain regions. Previously experienced or learned activity is believed to be replayed during sleep \cite{ji2007coordinated,wilson1994reactivation}.
This replay of the recently learned memories along with relevant old memories is thought to be the critical mechanism that results in memory consolidation.
In this new study, we implemented the main mechanisms behind the sleep neuronal activity to benefit ANNs performance based on the relevant biophysical modeling work \cite{krishnan2016cellular,hill2005modeling,bazhenov2002model}.

The principles of memory consolidation during sleep have previously been used to address the problem of catastrophic forgetting in ANNs. A generative model of the hippocampus and cortex was used to generate examples from a distribution of previously learned tasks in order to retrain (replay) these tasks during an off-line phase \cite{kemker2017fearnet}. Generative algorithms were used to generate previously experienced stimuli during the next training period in \cite{van2018generative, li2018learning}. A loss function (termed elastic weight consolidation - EWC), which penalizes updates to weights deemed important for previous tasks, was introduced in \cite{kirkpatrick2017overcoming} making use of synaptic mechanisms of memory consolidation. Although these studies report positive results in preventing catastrophic forgetting, they have many limitations. First, EWC does not seem to work in an incremental learning framework \cite{van2018generative, kemker2018measuring}. Second, generative models only focus on the replay aspect of sleep and therefore it is unclear if these models could have any benefits in addressing problems of generalization of knowledge. Further, generative models require a separate network that stores the statistics of the previously learned inputs which imposes an additional cost, while rehearsal of small examples of different classes may be sufficient to prevent catastrophic forgetting \cite{Hayes2018Memoryb}.

In this work, we propose a novel sleep algorithm which makes use of two principles observed during sleep in biology: memory reactivation and synaptic plasticity. First, we train ANN using backpropagation algorithm. After initial training, denoted awake training, we convert the ANN to SNN and perform unsupervised STDP phase with noisy input and increased intrinsic network activity to simulate sleep-like active (Up) state dynamics found during deep sleep. Finally, the weights from the SNN are converted back to the ANN and we test performance. We uncover three benefits of using this sleep algorithm. 
\begin{enumerate}
\itemsep0em 
  \item Sleep reduces catastrophic forgetting by reactivation of the older tasks.
  \item Sleep increases the network's ability to generalize to noisy or alternated versions of the training data set.
  \item Sleep allows the network to perform forward transfer learning.
\end{enumerate}

To the best of our knowledge, this is the first known sleep-like algorithm that improves ANNs ability to generalize on the noisy or alternated versions of the input. While few other algorithms were previously proposed to prevent catastrophic forgetting \cite{van2018generative, Hayes2018Memoryb, kemker2017fearnet}, our approach is more scalable and it does not require storage of the previously seen inputs or using pseudo-rehearsal to regenerate and retrain those inputs.  Importantly we demonstrate that ANNs retain information about (what seems to be) forgotten tasks that could be recovered during sleep. Our algorithm can be complimentary to the other approaches and, importantly, it provides a principled way to incorporate various features of sleep to the wide range of neural network architectures. 

\section{Methods} 
\label{methods}

First, we describe the general components of the sleep algorithm. Briefly, a fully connected feedforward network (FCN) is trained on a task. The ANN consisted of ReLU activation units to create positive firing rates and no bias. We used a previously developed algorithm to convert the architecture in the FCN to an equivalent SNN \cite{diehl2015fast}. In short, the weights are transferred directly to the SNN, which consists of leaky integrate and fire neurons. Weights are scaled by the maximum activation in each layer during training. After building the SNN, we run a 'sleep' phase which modifies the network connectivity based on spike-timing dependent plasticity (STDP). After running sleep phase, the weights are converted back into the FCN and testing or further training is performed.

Below, we describe the sleep phase in more details. The input layer of the SNN is activated with  Poisson-distributed spike trains with mean firing rate given by the average value of each unit activation in the ANN for all tasks seen so far (during initial training). We presented either the entire average image seen during initial ANN training or randomized portions of the average image seen so far or all the active regions during any of the inputs. To apply STDP, we ran one time step of the network propagating activity. Each layer in the SNN is characterized by two important parameters that dictates its firing rate: a threshold and a synaptic scaling factor. The input to a neuron is computed as $a\mathbf{W} \dot x$, where $a$ is the layer-specific synaptic scaling factor, $\mathbf{W}$ is the weight matrix, and $x$ is the spiking activity (binary) of the previous layer. This input is added to the neuron's membrane potential. If the membrane potential exceeds a threshold, the neuron fires a spike and its membrane potential is reset. Otherwise, the potential decays exponentially. After each spike, weights are updated according to a modified sigmoidal weight-dependent STDP rule. Weights are increased if a pre-synaptic spike leads to a post-synaptic spike. Weights are decreased if a post-synaptic spike fires without a pre-synaptic spike.

We tested the sleep algorithm on various datasets, including toy datasets which was used as a motivating example. This dataset, termed "Patches", consists of 4 images of binary pixels arranged in an $N\times N$ matrix. Each of the images has varying amount of overlap with the other 4 images to test the catastrophic forgetting. Likewise, we blurred the patches so that on-pixels spillover into the neighboring pixels making the dataset slightly different from the one the network was trained on. We used this dataset to show the benefits of the sleep algorithm in a simpler setting. We also tested the sleep algorithm on the MNIST \cite{lecun1998gradient} and CUB200 \cite{welinder2010caltech} datasets to ensure generalizability of our approach. For CUB200, we used the pre-trained Resnet embeddings previously used for catastrophic forgetting \cite{kemker2017fearnet, he2016deep}.

To test catastrophic forgetting, we utilized an incremental learning framework. The FCN was trained sequentially on groups of 2 classes for patches and MNIST and groups of 100 classes for CUB200 \cite{van2018generative}. After training on a single task, we run the sleep algorithm as described above before training on the next task. To test generalization, we trained FCN on the entire dataset and we compared this network's performance on classifying noisy or blurred images to the FCN performance that implemented sleep phase after training. For transfer learning, a network trained on one task was put to sleep and then tested on a new, unseen task. Dataset specific parameters for training and sleep in the catastrophic forgetting task are shown in Table \ref{table1}. For the MNIST dataset, we utilized a genetic algorithm to find optimal parameters, although this is not an absolute requirement and our summary results are based on hand-tuned parameters.

\begin{table}[H]
\begin{tabular}{|l|l|l|l|}
\hline
 & Patches & MNIST & CUB200 \\ \hline
Architecture & {[}100, 4{]} & {[}784, 1200, 1200, 10{]} & {[}2048, 350, 300, 200{]} \\ \hline
Learning Rate & 0.1 & 0.065 & 0.1, 0.01 \\ \hline
Dropout & 0 & 0.2 & 0.25 \\ \hline
Epochs & 1 per task & 2 per task & 50 per task \\ \hline
Input Rate & 64 & 130 Hz & 32 \\ \hline
Thresholds & 1.045 & 2.1772, 1.5217,  0.9599 & 1, 1, 1 \\ \hline
Synaptic & 4.25 & 3.4723, 25.52, 2.4186 & 1, 1, 1 \\ \hline
Increase factor & 0.0035 & 0.0197 & 0.01 \\ \hline
Decrease factor & 0.0002 & 0.0016 & 0.001 \\ \hline
\end{tabular}
\caption{Approximate description of parameters used in each of the 3 datasets.}
\label{table1}
\end{table}

\section{Results}
\label{results}


\subsection{Sleep prevents catestrophic forgetting and lead to forward transfer for Patches}

\begin{figure}[t]
  \centering
  \includegraphics[width=5.5in,height=8cm]{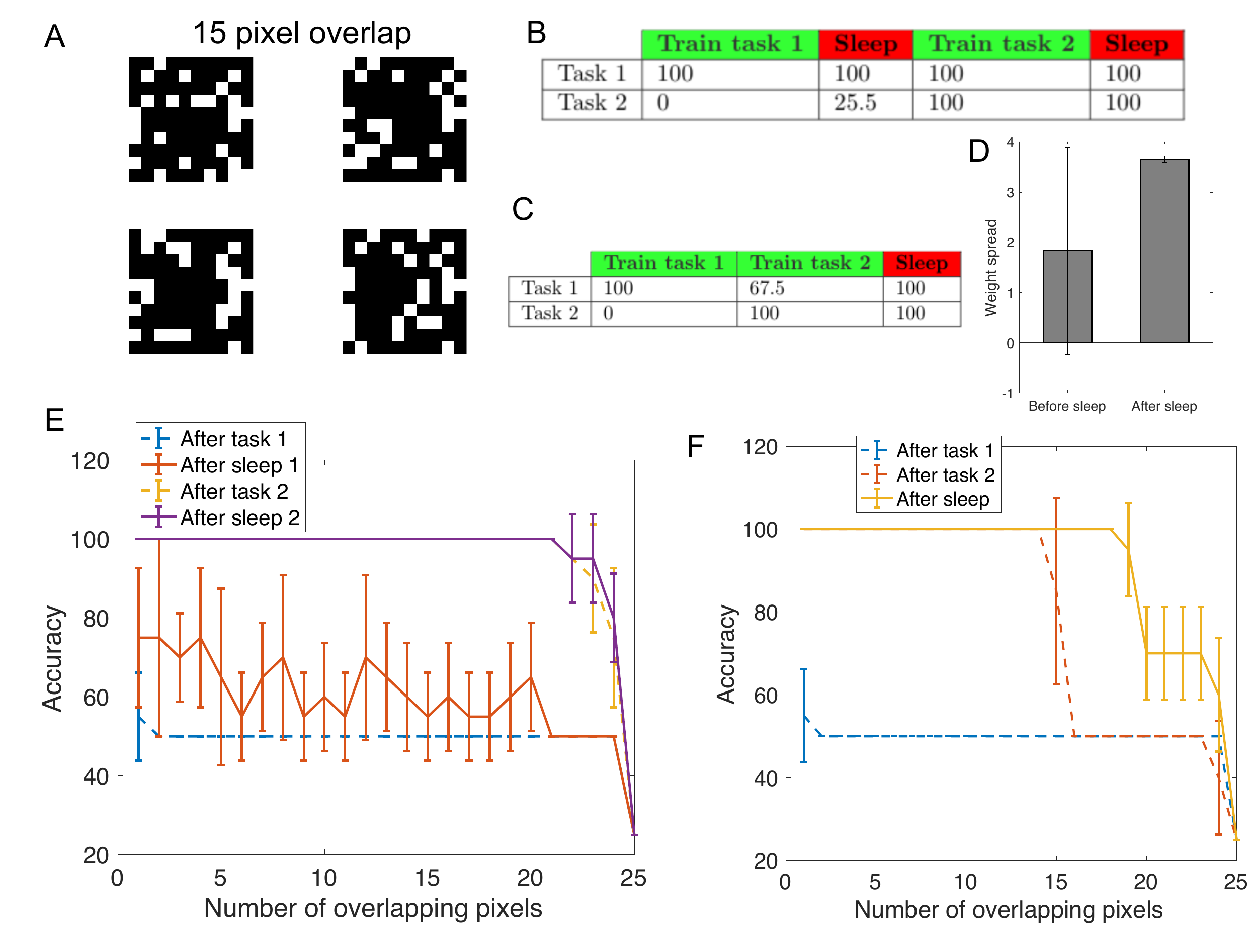}
  \caption{Sleep reduces catastrophic forgetting and increases forward learning in Patches dataset. A) Example of the Patches dataset with 4 images with 15 pixel overlap among the images. B) Accuracy over 100 trials on task 1 (first 2 images) and task 2 (second 2 images) after training on task 1, a first sleep phase, training on task 2, and a second sleep phase. C) Same as B with only one final sleep phase. D) Spread of the weights of weights connecting from on-pixels to output neurons vs. off-pixels. E) Accuracy as a function of number of overlapping pixels at different points in training (blue dashed = after task 1, red - after first sleep, yellow dashed - after task 2, purple - after final sleep, 5 trials) F) Same as D but with one final sleep phase indicating that intermediate sleep helps forward learning.}
  \label{fig2}
\end{figure}

The Patches dataset represents an easily interpretable example to verify and validate our sleep algorithm. We utilized 4 binary images of size $10\times 10$ with 15 pixel overlap and 25\% of pixels turned on. Thus, 10 pixels are unique for each image in the dataset (Fig. \ref{fig2}A). To determine if catastrophic forgetting occurs in this model, and if sleep can recover performance, we split the dataset into two tasks - task one  representing two  images (out of four total) and the other task comprised of the other two images. Training on task 1 resulted in the high performance on task 1 with no performance improvement on task 2. After sleep phase, performance on task 1 remained perfect, while task 2 performance sometimes revealed an increase. After training on task 2, performance on task 1 on average decreased from its perfect level, indicating forgetting of task 1. However, after sleep, performance on both task 1 and task 2 reached 100\% (Fig \ref{fig2}B). Including only one sleep phase at the end of awake training also resurrected performance for both tasks (Fig. \ref{fig2}C). 

To analyze why sleep prevents catastrophic forgetting in this toy example, we looked at the weights connecting to each input neuron. Since we have knowledge of all the pixels in the training data, we could measure the weights connecting from the pixels that are turned on during image presentation to the corresponding output neuron. Ideally, for a given image, the spread between weights from on-pixels and weights from off-pixels should be high, such that on-pixels drive an output neuron and off-pixels suppress the same output neuron. To measure this, we computed the average spread across output neurons and weights for on-pixels and off-pixels (Fig. \ref{fig2}D). Our results indicate that sleep increases the spread between weights originating from on-pixels vs those from off-pixels, validating that the sleep algorithm is acting by increasing meaningful weights and decreasing potentially irrelevant or incorrect weights.  
We next observed the performance as a function of the number of overlapping pixels in the dataset for 2 cases: one with sleep implementing after each awake training period and one with only one sleep phase at the end of training. With 2 sleep phases, we observed that after the first sleep episode, the network performed well on the first task and  correctly classified images from the second task about 50\% of the time (Fig. \ref{fig2}E). This suggests that sleep may lead to increase in performance on tasks for which  SNN has not seen any training data inputs. We call such an improvement on the previously unseen  tasks as 'forward transfer' similar to zero-shot learning phenomenon previously shown in other architectures, e.g. \cite{palatucci2009zero,socher2013zero}. 

After training on the second task followed by sleep, the network classified all the images correctly up to the very high level of the pixel overlap. In the later case, we observed that the sleep phase increases performance beyond that of the control network, indicating less catastrophic forgetting (Fig. \ref{fig2}F). Forgetting only occurs for pixel overlap greater than 15 pixels. However, for higher pixel overlap values, sleep routinely reduced the amount of forgetting. Comparing the two cases, we note that an intermediate sleep phase between task one and task two actually increases performance and reduces forgetting after normal awake training on task two. This again suggests that sleep may be useful in creating a forward transfer representation of similar, yet novel, tasks and may boost transfer learning in other domains. Overall, these results provide validation of our sleep algorithm and raise the question if the same results can be obtained for more complex datasets and network architectures, which we will discuss later in this paper.

\subsection{What causes catastrophic forgetting and how does sleep help?}
\label{analysis}
In this section, we consider a simple case study to examine the cause of the catastrophic failure and the role of sleep in recovering from the forgetting. While this example is not intended to model all scenarios of catastrophic forgetting, it extracts the intuition and explains the basic mechanism behind our algorithm. \\
\noindent
Let us consider the 3-layer network trained on two categories, each with just one example. Consider 2 binary vectors (Category 1 and Category 2) with some region of overlap.\\ 
We consider ReLU activation since it was used in the rest of this work. We assume the output to be the neuron with the highest activation in the output layer. Let the network be trained on Category 1 with backpropagation using static learning rate. Following this, we trained the network on Category 2 using same approach. A 3-layer network we consider here has an input layer with $10$ neurons, $30$ hidden neuron and an output layer with $2$ neurons for the $2$ categories.  Inputs are 10 bits long with 5 bit overlap. We trained with learning rate of 0.1 for 4 epochs. 

\textit{Analysis of hidden layer behaviour:}
We can divide the hidden neurons into four types based on their activation for the two categories: $A$ - those neurons that fire for Category 1 but not 2; $B$ - those neurons that fire for Category 2 but not 1; $C$ - those neurons that fire for Category 1 and 2; $D$ - those that fire for neither category, where firing indicates a non-zero activation. Note that these sets may change during training or sleep. Let $X_i$ be the weights from type $X$ to output $i$.\\
Consider the case where the input of Category 1 is presented. The only hidden layer neurons that fire are $A$ and $C$. Output neuron 1 will get the net value $A*A_1 + C*C_1$ and output neuron 2 will get the net value $A*A_2 + C*C_2$. For output neuron 1 to fire, we need two conditions to be held: (1) $A*A_1 + C*C_1 > 0 $ (2) $A*A_1 + C*C_1 > A*A_2 + C*C_2$. 
The second condition above can be rewritten as $A*A_2 - A*A_1 < C*C_1 - C*C_2$, which separates the weights according to the hidden neurons. Using this separation, we give the following definitions:
Define $a$ to be $(A_2-A_1)*A$ on pattern $1$; $b$ to be $(A_2-A_1)*A$ on pattern $2$; $p$ to be $(C_1-C_2)*C$ on pattern $1$ and $q$ to be $(C_1-C_2)*C$ on pattern $2$. (Note that $p$ and $q$ are very closely correlated since they differ only in the activation values of $C$ neurons which are positive in both cases). \\
So, on the input pattern $1$, output 1 fires only if $a < p$; on input pattern $2$, output 2 fires only if $q < b$.

\textit{Catastrophic forgetting:}
Following training on 2 categories, if the network can not recall Category 1, i.e., output neuron 1 activation is negative or less than that of output neuron 2, catastrophic forgetting has occurred (We confirmed this occurred 78\% of times for the 3 layer network described above and 100 trials).
The second phase of training ensures $q < b$. This could involve reduction in $q$ which would reduce $p$ as well.  (Since $A$ does not fire on input pattern 2, back-propagation does not alter $a$)  Reducing $p$ may result in failing the condition $a < p$, i.e., misclassifying input 1.

\textit{Effect of sleep:}  Sleep can increase  difference among weights (which are different enough to begin with) as was shown in  \cite{Wei2017Differential, gonzalez2019can}. So, as the difference between $A_2$ and $A_1$ increases, this decreases $a$ (as $A_1$ is higher, $a = A_2-A_1$ decreases). Occurrence of the same change to $p$ is prevented as follows: it is likely that at least one of the weights coming to a $C$ neuron is negative. In that case, increasing the difference would involve making the negative weight even more negative,  resulting in the neuron  joining either $A$ or $B$ (as it no longer fires for the pattern showing the negative weight), thus reducing $p$. (This is explained further in the supplement)\\
When the neurons in $C$ remains, we have a more complex case: here, $a$ decreases, but $p$ may also decrease correspondingly; another undesirable scenario is when  $b$ decreases to become less than $q$. Typically sleep tends to drive synaptic weights of the opposite signs, or the weights of same sign but different by some threshold value, away from each other. There are conditions when the difference between weights is below threshold point to cause divergence. In those cases sleep does not improve performance.

\textit{Experiments:} In our experiments, for majority of the cases, we found $C$ to be empty after sleep, thus making $p$ to become $0$. For the instances when this was not a case,  the initial values of $A_1$, $A_2$, $B_1$ and $B_2$ were almost$0$, i.e., the entire work of classifying the inputs is done by shared input. In such case, the network has no hidden information that sleep could retrieve. (Evidence is provided in the supplement). 

\subsection{Sleep recovers tasks lost due to catastrophic forgetting in MNIST and CUB200}
\begin{figure}[H]
  \centering
  \includegraphics[width=\textwidth]{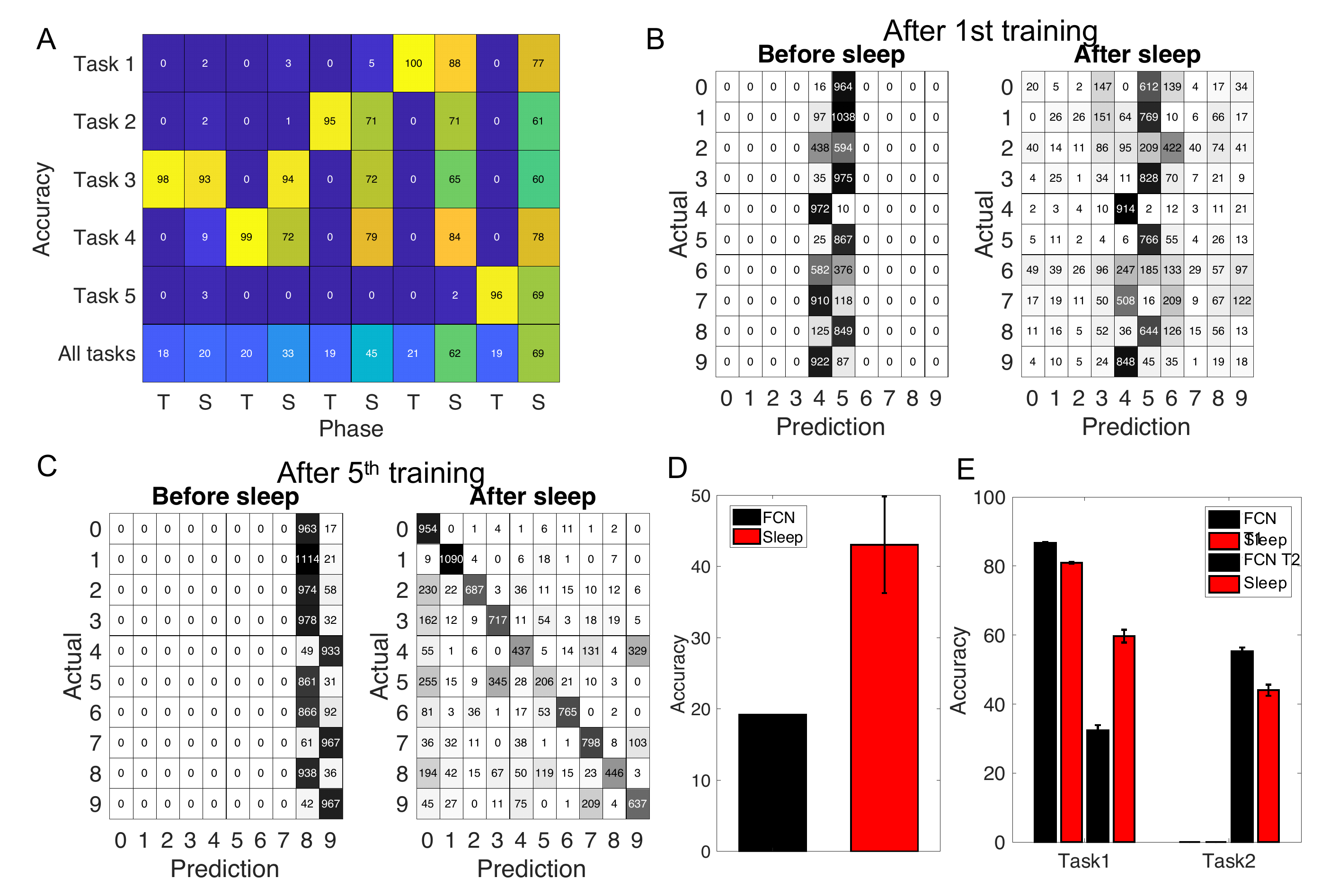}
  \caption{Sleep reduces catastrophic forgetting in MNIST and CUB200 datasets. A) Rounded accuracy for each of the 5 tasks (first 5 rows) and overall (6th row) as a function of training phase - T - awake training, S - Sleep. B) Confusion matrix after the first awake and sleep phase shows some forward zero-shot learning C) Same as B but after last training and sleep phase. D) Summary MNIST performance with sleep (red) vs. a simple fully connected network (black) averaged after different task orders. E) Accuracy for task 1 (left group of bars) and task 2 (right group) after training on task 1 (first black bar), first sleep phase (first red bar), training on task 2 (second black bar), and second sleep phase (last red bar) for CUB200.}
  \label{fig3}
\end{figure}

ANNs have been shown to suffer from catastrophic forgetting whereby they perform well on the recently learned tasks but fail at previously learned tasks for various datasets, including MNIST and CUB200 \cite{kemker2018measuring}. Here, we created 5 tasks for the MNIST dataset and 2 tasks for the CUB200 dataset. Each pair of digits in MNIST was defined as a single task, and half of the classes in CUB200 was considered to be a single task. Each task was incrementally trained, followed by a sleep phase, until all tasks were trained. A baseline network trained incrementally without sleep performed poorly (Fig. \ref{fig3}D, black bar). However, we noted a significant improvement in the overall performance, as well as task specific performance, when sleep algorithm was incorporated into the training cycle (Fig. \ref{fig3}D, red bar). 

For MNIST, we found that each of the five tasks revealed an increase in classification accuracy after sleep even after being completely "forgotten" during awake training (Fig. \ref{fig3}A). For the 1st training + sleep cycle, the "before sleep" network only classifies images for the task that was seen during last training (digits 4-5 in Fig. \ref{fig3}B). After sleep, performance remains high on digits 4 and 5 but there is also spillover into the other digits. For the last training + sleep cycle, we observed the same effect. Only last task performed well right after the training (Fig. \ref{fig3}C). After sleep, performance on almost all digits nearly recovered (Fig. \ref{fig3}D). On the CUB200 dataset, we found that sleep can recover task 1 performance after training on task 2, with only minimal loss to task 2 performance (Fig. \ref{fig3}E). In conclusion, the sleep algorithm reduces catastrophic forgetting by reducing overlap between network activity for distinct classes.

Although specific performance numbers we obtain here are not as impressive as for some generative models \cite{van2018generative, kemker2017fearnet}, they surpass certain regularization methods, such as EWC, on incremental learning \cite{van2018generative}. Overall, we believe that the sleep algorithm can reduce catastrophic forgetting and interference with very little knowledge of the previously learned examples solely by utilizing STDP to reactivate forgotten weights. Ultimately, these results suggest that information about old tasks is not completely lost when catastrophic forgetting occurs from performance level perspective. Instead, information about old tasks remains present in the connectivity weights and offline STDP phase can resurrect this hidden information. To achieve higher performance,  offline STDP/sleep algorithm could be combined with generative replay to utilize specific, rather than average (as we use in our study here), inputs during sleep.

\subsection{Sleep promotes separation of internal representations for different inputs}
As suggested by our previous analysis \hyperref[sec:analysis]{section}, sleep could separate the neurons belonging to the different input categories and prevent catastrophic forgetting. This would also result in a change in the internal representation of the different inputs in the network. We examined this in the network trained on MNIST dataset and we compared performance before and after the sleep. 
\begin{figure}[t!]
  \centering
  \includegraphics[width=\textwidth]{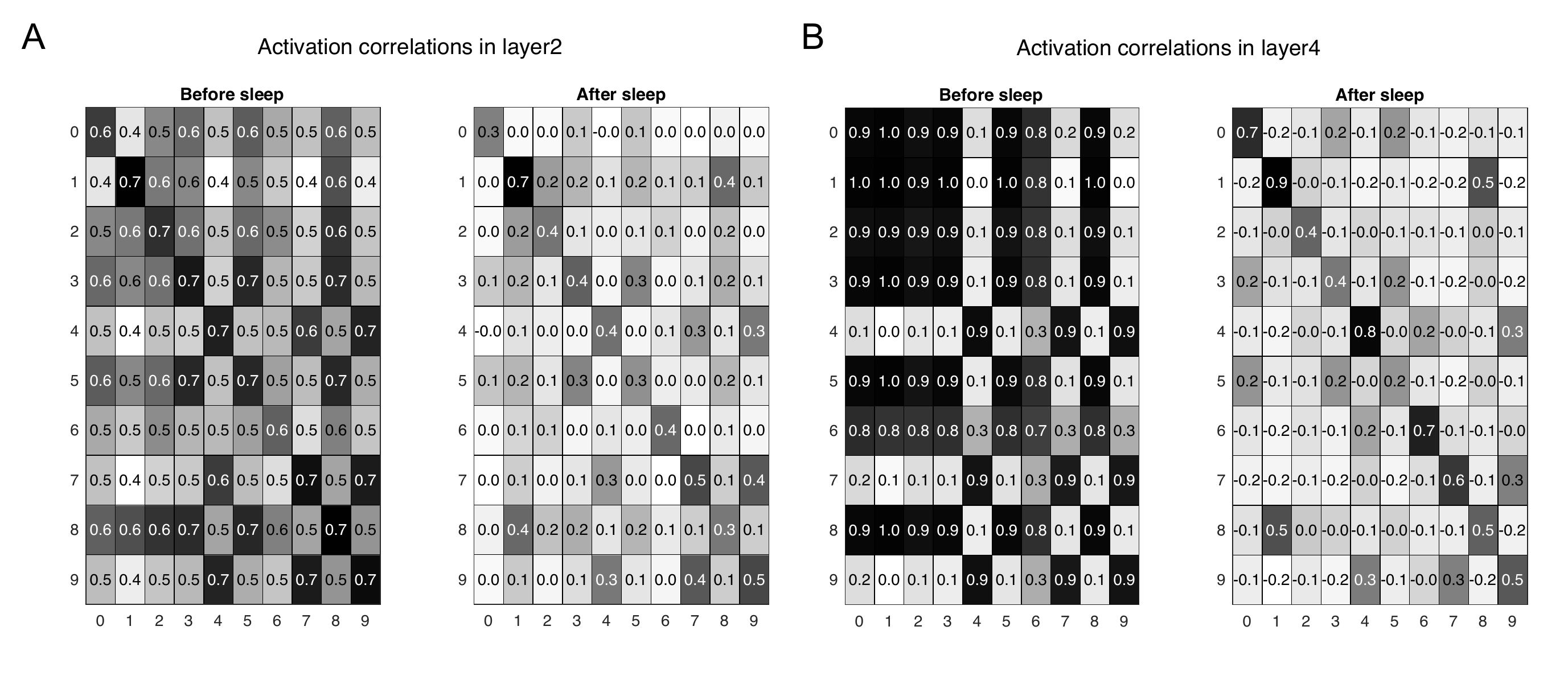}
  \caption{Sleep decreases representational overlap between MNIST classes at all layers. A) Average correlations of activations in the first hidden layer for each digit, i.e. the number in row 0 and column 5 indicates the average correlation of the activations of all examples of digit 0 and all examples of digit 5. B) Same as A except correlations are computed in the output layer.}
  \label{fig1}
\end{figure}
In order to examine how the internal representation of the different tasks are related and modified after sleep, we examined the correlation between ANN activation at different layers after awake training and after sleep. In particularly, we computed the average correlation between activation of examples of the class $i$ with examples of the class $j$.
We observed the correlation before sleep was higher both within the same input category and across all categories. After sleep, the correlations between different categories were reduced (Fig. \ref{fig1}) while the correlation within category remained high. This proposes that sleep promotes decorrelating the internal representations of the input categories, suggesting a mechanism by which sleep can prevent catastrophic forgetting.

\subsection{Sleep improves generalization}
\begin{figure}[t!]
  \centering
  \includegraphics[width=5.5in, height=8cm]{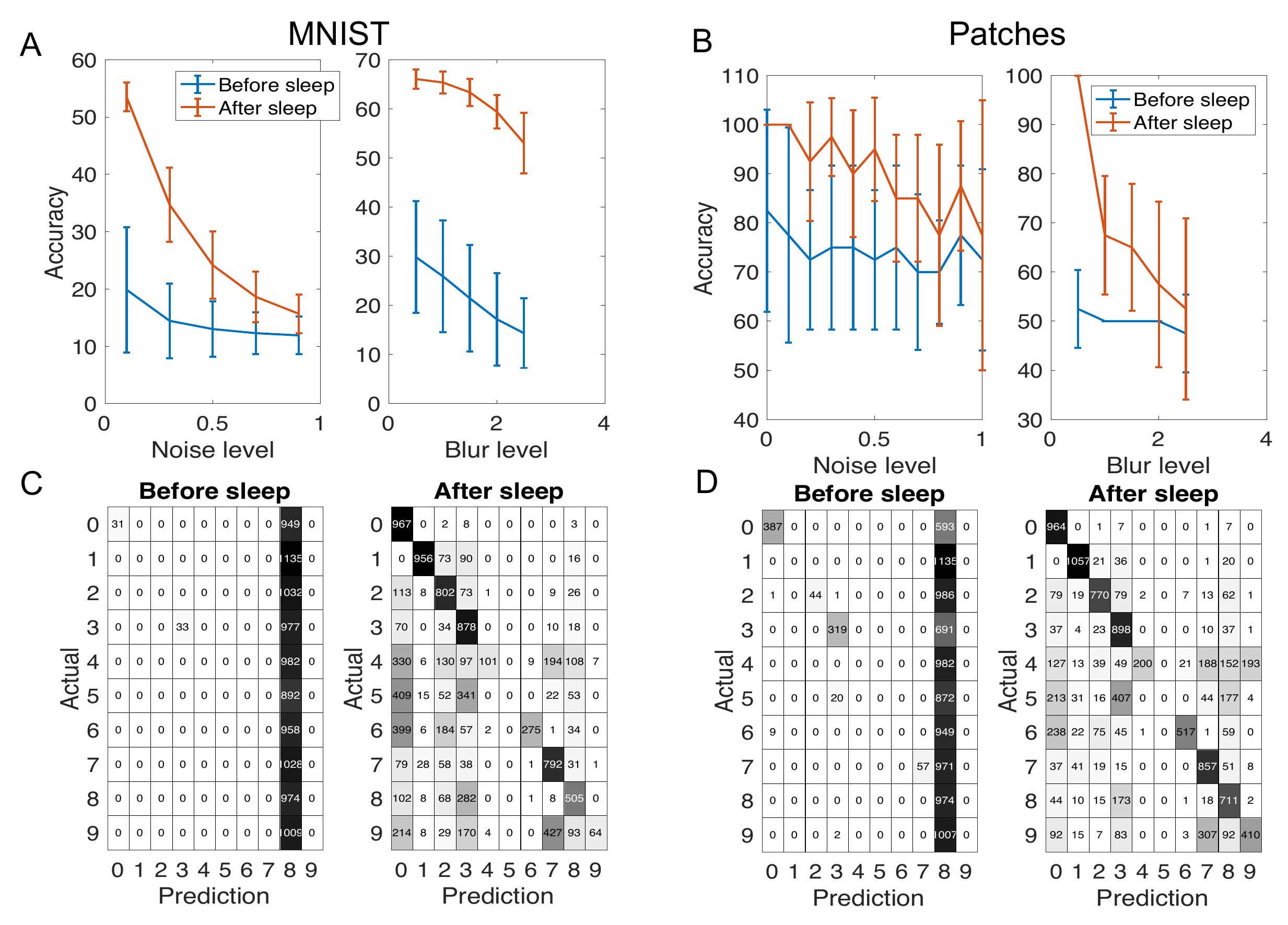}
  \caption{Sleep increases generalization performance on MNIST and Patches task.  A) A sub-optimal network is test on Gaussian noise (left) and Gaussian blurring (right) with sigma given by the blur level. Accuracy is shown as a function of degradation level after applying sleep (red) to before sleep (blue) averaged over 5 trials. B) Same as A for the Patches example. C-D) Confusion matrix before and after sleep for low noise and blur, respectively. See supplement for example images.}
  \label{fig4}
\end{figure}

Many studies in machine learning reported a failure of neural networks to generalize beyond their explicit training set \cite{geirhos2018generalisation}. Given that sleep tends to create a more generalized representation of the stimulus within network architecture, next we tested the hypothesis that sleep algorithm could increase ANN's ability to generalize beyond the original training data set. To do so, we created noisy and blurred versions of the MNIST and Patches samples and we tested the network before and after sleep on these distorted datasets (Fig. \ref{fig4}). Our results suggest that sleep can substantially increase the network's ability to classify degraded images. Indeed, for both  MNIST and Patches dataset, the "after sleep" network substantially outperformed the "before sleep" network on classifying noisy and blurred images. This is illustrated by  analysis of the confusion matrices, where "before sleep" network trained on the intact MNIST images favors one class over another when tested on the degraded images. Surprisingly, sleep restored the network ability to correctly predict the classes. It is important to note that we trained  MNIST network sub-optimally to illustrate the case where the network performs low on degraded images. The same network architecture can perform well without sleep even on degraded images if the training dataset is significantly expanded.

These results highlight the benefit of utilizing sleep to generalize representation of the task at hand. ANNs are normally trained on the highly filtered datasets that are identically and independently distributed. However, in a real-world scenario, inputs may not meet these assumptions. Incorporating a sleep-like phase into training of ANNs may enable a more generalized representation of the input statistics, such that distributions which are not explicitly trained may still be represented by the network after sleep. 

\section{Discussion}
We showed that a biologically-inspired sleep algorithm may provide several important benefits when incorporated to the neural network training. We found that sleep is able to resurrect tasks that were erased due to the catastrophic forgetting after new task training utilizing backpropagation algorithm. Our study suggests that while performance on such "forgotten" tasks was dramatically reduced after new training, the network weights retained partial information about the older tasks and sleep could reactivate the older tasks to strengthen the reduced connectivity and to recover the  performance. 

While proposed sleep method to prevent catastrophic forgetting currently performs below some other techniques \cite{van2018generative,kemker2017fearnet,Hayes2018Memoryb}, the other approaches either remember full set of training inputs or recreate inputs from the generator networks. Our approach does not require storing any  input information and it may be complimentary to the other techniques; in that applying a sleep-like phase to the generative mechanisms may further boost overall performance.


We found that sleep algorithm can also help to generalize on the previously learning tasks. Indeed, classification accuracy increased significantly after sleep for images that incorporated Gaussian noise or were blurred. We used MNIST dataset to demonstrate this effect which improved performance from about 20\% to 50\%. 
This additional benefit of sleep likely arises from stochastic nature of the network dynamics during sleep that creates a more generalized representation of the previously learned tasks. Indeed, we found (not shown) that the same approach can be extended to increase network resistance to adversarial attacks. 

Finally, we also observed that sleep improves performance on the tasks that the network has not been trained on, but that share some properties with the previously trained tasks. We refer to this effect as 'forward transfer', similar to zero-shot learning \cite{palatucci2009zero, socher2013zero}. This effect again likely arises from the stochasticity of the sleep dynamics which allows for the shared features between tasks to be strengthened which are then used in the backpropagation phase to learn different tasks.

There are several current and past attempts to implement  effect of sleep in ANNs or machine learning architectures \cite{hinton1995wake,kemker2017fearnet}. However, our approach significantly differs from these previous attempts, in that we used conversion method from ANN to SNN and implemented sleep at the SNN level which is relatively well understood from the neuroscience perspective \cite{krishnan2016cellular,hill2005modeling,wei2016synaptic,Wei2017Differential}. Importantly, this approach allows direct implementation of the many other brain inspired ideas. To sum, we believe that our approach provides a principled way to apply mechanisms of the biological sleep in memory consolidation to existing AI architectures.

\medskip
\small
\bibliography{references} 

\begin{thebibliography}{10}

\bibitem{russakovsky2015imagenet}
Olga Russakovsky, Jia Deng, Hao Su, Jonathan Krause, Sanjeev Satheesh, Sean Ma,
  Zhiheng Huang, Andrej Karpathy, Aditya Khosla, Michael Bernstein, et~al.
\newblock Imagenet large scale visual recognition challenge.
\newblock {\em International journal of computer vision}, 115(3):211--252,
  2015.

\bibitem{silver2018general}
David Silver, Thomas Hubert, Julian Schrittwieser, Ioannis Antonoglou, Matthew
  Lai, Arthur Guez, Marc Lanctot, Laurent Sifre, Dharshan Kumaran, Thore
  Graepel, et~al.
\newblock A general reinforcement learning algorithm that masters chess, shogi,
  and go through self-play.
\newblock {\em Science}, 362(6419):1140--1144, 2018.

\bibitem{goodfellow2013empirical}
Ian~J Goodfellow, Mehdi Mirza, Da~Xiao, Aaron Courville, and Yoshua Bengio.
\newblock An empirical investigation of catastrophic forgetting in
  gradient-based neural networks.
\newblock {\em arXiv preprint arXiv:1312.6211}, 2013.

\bibitem{mcclelland1995there}
James~L McClelland, Bruce~L McNaughton, and Randall~C O'reilly.
\newblock Why there are complementary learning systems in the hippocampus and
  neocortex: insights from the successes and failures of connectionist models
  of learning and memory.
\newblock {\em Psychological review}, 102(3):419, 1995.

\bibitem{geirhos2018generalisation}
Robert Geirhos, Carlos~RM Temme, Jonas Rauber, Heiko~H Sch{\"u}tt, Matthias
  Bethge, and Felix~A Wichmann.
\newblock Generalisation in humans and deep neural networks.
\newblock In {\em Advances in Neural Information Processing Systems}, pages
  7538--7550, 2018.

\bibitem{dodge2017study}
Samuel Dodge and Lina Karam.
\newblock A study and comparison of human and deep learning recognition
  performance under visual distortions.
\newblock In {\em 2017 26th international conference on computer communication
  and networks (ICCCN)}, pages 1--7. IEEE, 2017.

\bibitem{szegedy2013intriguing}
Christian Szegedy, Wojciech Zaremba, Ilya Sutskever, Joan Bruna, Dumitru Erhan,
  Ian Goodfellow, and Rob Fergus.
\newblock Intriguing properties of neural networks.
\newblock {\em arXiv preprint arXiv:1312.6199}, 2013.

\bibitem{pan2009survey}
Sinno~Jialin Pan and Qiang Yang.
\newblock A survey on transfer learning.
\newblock {\em IEEE Transactions on knowledge and data engineering},
  22(10):1345--1359, 2009.

\bibitem{spengler1997learning}
Friederike Spengler, Timothy~PL Roberts, David Poeppel, Nancy Byl, Xiaoqin
  Wang, Howard~A Rowley, and Mike~M Merzenich.
\newblock Learning transfer and neuronal plasticity in humans trained in
  tactile discrimination.
\newblock {\em Neuroscience letters}, 232(3):151--154, 1997.

\bibitem{walker2004sleep}
Matthew~P Walker and Robert Stickgold.
\newblock Sleep-dependent learning and memory consolidation.
\newblock {\em Neuron}, 44(1):121--133, 2004.

\bibitem{stickgold2013sleep}
Robert Stickgold and Matthew~P Walker.
\newblock Sleep-dependent memory triage: evolving generalization through
  selective processing.
\newblock {\em Nature neuroscience}, 16(2):139, 2013.

\bibitem{rasch2013sleep}
Bj{\"o}rn Rasch and Jan Born.
\newblock About sleep's role in memory.
\newblock {\em Physiological reviews}, 93(2):681--766, 2013.

\bibitem{ji2007coordinated}
Daoyun Ji and Matthew~A Wilson.
\newblock Coordinated memory replay in the visual cortex and hippocampus during
  sleep.
\newblock {\em Nature neuroscience}, 10(1):100, 2007.

\bibitem{wilson1994reactivation}
Matthew~A Wilson and Bruce~L McNaughton.
\newblock Reactivation of hippocampal ensemble memories during sleep.
\newblock {\em Science}, 265(5172):676--679, 1994.

\bibitem{krishnan2016cellular}
Giri~P Krishnan, Sylvain Chauvette, Isaac Shamie, Sara Soltani, Igor Timofeev,
  Sydney~S Cash, Eric Halgren, and Maxim Bazhenov.
\newblock Cellular and {{Neurochemical Basis}} of {{Sleep Stages}} in the
  {{Thalamocortical Network}}.
\newblock 5:e18607.

\bibitem{hill2005modeling}
Sean Hill and Giulio Tononi.
\newblock Modeling {{Sleep}} and {{Wakefulness}} in the {{Thalamocortical
  System}}.
\newblock 93(3):1671--1698.

\bibitem{bazhenov2002model}
Maxim Bazhenov, Igor Timofeev, Mircea Steriade, and Terrence~J Sejnowski.
\newblock Model of thalamocortical slow-wave sleep oscillations and transitions
  to activated states.
\newblock {\em Journal of neuroscience}, 22(19):8691--8704, 2002.

\bibitem{kemker2017fearnet}
Ronald Kemker and Christopher Kanan.
\newblock Fearnet: Brain-inspired model for incremental learning.
\newblock {\em arXiv preprint arXiv:1711.10563}, 2017.

\bibitem{van2018generative}
Gido~M van~de Ven and Andreas~S Tolias.
\newblock Generative replay with feedback connections as a general strategy for
  continual learning.
\newblock {\em arXiv preprint arXiv:1809.10635}, 2018.

\bibitem{li2018learning}
Zhizhong Li and Derek Hoiem.
\newblock Learning without forgetting.
\newblock {\em IEEE transactions on pattern analysis and machine intelligence},
  40(12):2935--2947, 2018.

\bibitem{kirkpatrick2017overcoming}
James Kirkpatrick, Razvan Pascanu, Neil Rabinowitz, Joel Veness, Guillaume
  Desjardins, Andrei~A Rusu, Kieran Milan, John Quan, Tiago Ramalho, Agnieszka
  Grabska-Barwinska, et~al.
\newblock Overcoming catastrophic forgetting in neural networks.
\newblock {\em Proceedings of the national academy of sciences},
  114(13):3521--3526, 2017.

\bibitem{kemker2018measuring}
Ronald Kemker, Marc McClure, Angelina Abitino, Tyler~L Hayes, and Christopher
  Kanan.
\newblock Measuring catastrophic forgetting in neural networks.
\newblock In {\em Thirty-second AAAI conference on artificial intelligence},
  2018.

\bibitem{Hayes2018Memoryb}
Tyler~L. Hayes, Nathan~D. Cahill, and Christopher Kanan.
\newblock Memory {{Efficient Experience Replay}} for {{Streaming Learning}}.

\bibitem{diehl2015fast}
Peter~U Diehl, Daniel Neil, Jonathan Binas, Matthew Cook, Shih-Chii Liu, and
  Michael Pfeiffer.
\newblock Fast-classifying, high-accuracy spiking deep networks through weight
  and threshold balancing.
\newblock In {\em 2015 International Joint Conference on Neural Networks
  (IJCNN)}, pages 1--8. IEEE, 2015.

\bibitem{lecun1998gradient}
Yann LeCun, L{\'e}on Bottou, Yoshua Bengio, Patrick Haffner, et~al.
\newblock Gradient-based learning applied to document recognition.
\newblock {\em Proceedings of the IEEE}, 86(11):2278--2324, 1998.

\bibitem{welinder2010caltech}
Peter Welinder, Steve Branson, Takeshi Mita, Catherine Wah, Florian Schroff,
  Serge Belongie, and Pietro Perona.
\newblock Caltech-ucsd birds 200.
\newblock 2010.

\bibitem{he2016deep}
Kaiming He, Xiangyu Zhang, Shaoqing Ren, and Jian Sun.
\newblock Deep residual learning for image recognition.
\newblock In {\em Proceedings of the IEEE conference on computer vision and
  pattern recognition}, pages 770--778, 2016.

\bibitem{palatucci2009zero}
Mark Palatucci, Dean Pomerleau, Geoffrey~E Hinton, and Tom~M Mitchell.
\newblock Zero-shot learning with semantic output codes.
\newblock In {\em Advances in neural information processing systems}, pages
  1410--1418, 2009.

\bibitem{socher2013zero}
Richard Socher, Milind Ganjoo, Christopher~D Manning, and Andrew Ng.
\newblock Zero-shot learning through cross-modal transfer.
\newblock In {\em Advances in neural information processing systems}, pages
  935--943, 2013.

\bibitem{Wei2017Differential}
Yina Wei, Giri~P. Krishnan, Maxim Komarov, and Maxim Bazhenov.
\newblock Differential roles of sleep spindles and sleep slow oscillations in
  memory consolidation.
\newblock page 153007.

\bibitem{gonzalez2019can}
Oscar~C Gonzalez, Yury Sokolov, Giri Krishnan, and Maxim Bazhenov.
\newblock Can sleep protect memories from catastrophic forgetting?
\newblock {\em BioRxiv}, page 569038, 2019.

\bibitem{hinton1995wake}
Geoffrey~E Hinton, Peter Dayan, Brendan~J Frey, and Radford~M Neal.
\newblock The" wake-sleep" algorithm for unsupervised neural networks.
\newblock {\em Science}, 268(5214):1158--1161, 1995.

\bibitem{wei2016synaptic}
Yina Wei, Giri~P Krishnan, and Maxim Bazhenov.
\newblock Synaptic {{Mechanisms}} of {{Memory Consolidation}} during {{Sleep
  Slow Oscillations}}.
\newblock {\em The Journal of Neuroscience}, 36(15):4231--4247, 2016.

\end{thebibliography}
\bibliographystyle{unsrt}



\section*{Supplementary material}

# 1 Supplementary Material

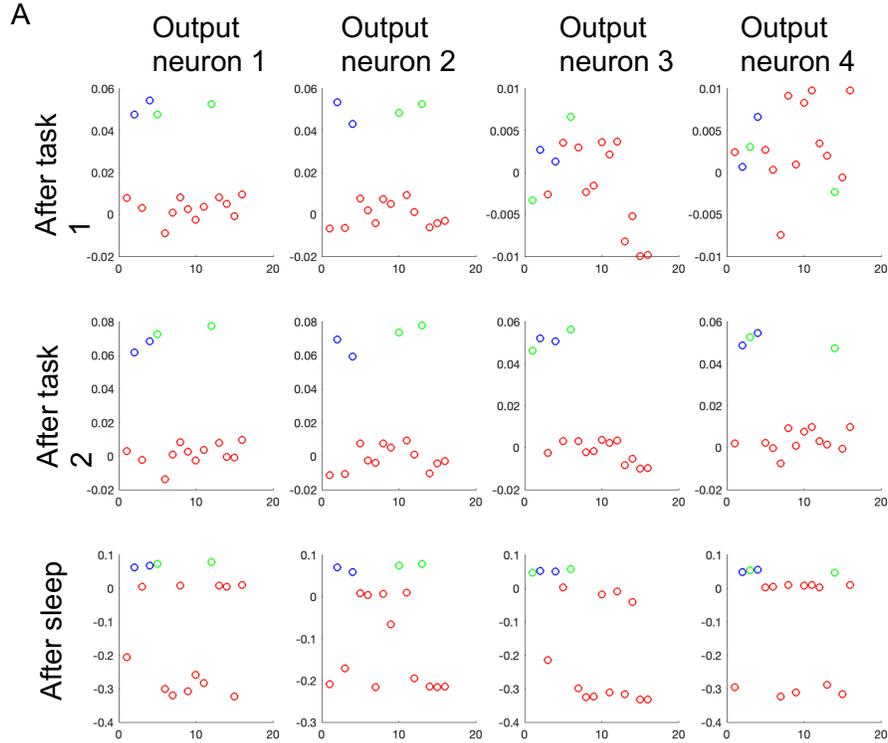

Figure 1: Weights connecting from image (4 × 4) to output neuron (columns) after task 1, after task 2, and after sleep (rows). Green pixels represent unique on-pixels for image $i$, where $i$ is the column number. Blue points are non-unique on-pixels. Red points are off-pixels. Y-axis is the value of the weights, and X-axis is pixel location in the image, representing each of the 16 pixel locations. This shows that sleep decreases the value of incorrect weights while maintaining and sometimes increasing value of positively identifying weights.

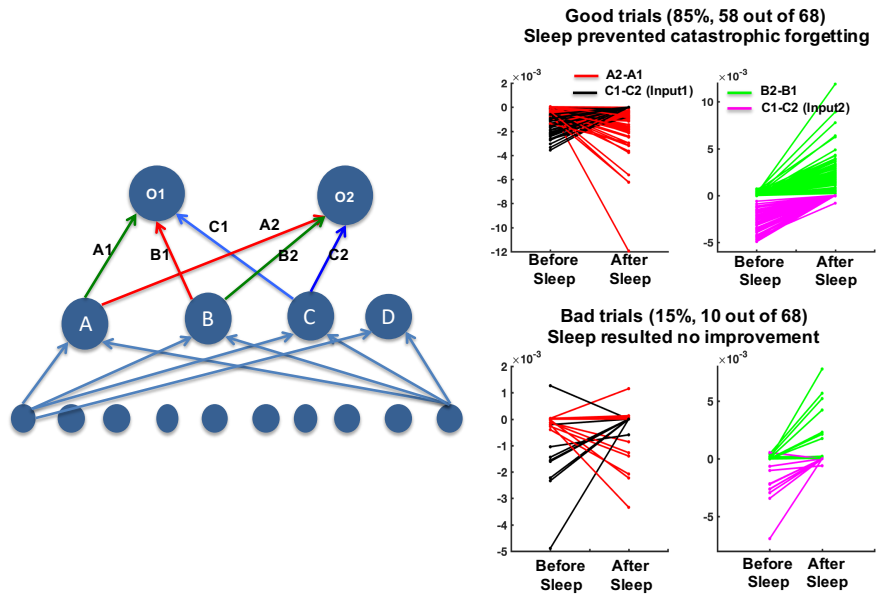

Figure 2: Example of the binary vector analysis. In the left graph, we show the structure of the network. A - fires only for input 1. B - fires only for input 2. C - fires for both inputs. D - fires for neither input 1 nor input 2. Green arrows represent desirable connections and red arrows indicate incorrect connections. Blue arrows are mixed depending on the input. The equations on the graphs on the right compare the difference between green and red arrows to the difference between blue arrows (for a given input). Depleting the set of C neurons correspond to giving the differences in the inputs more importance.

| | T1 | S | T2 | S | T3 | S | T4 | S | T5 | S |
|---|---|---|---|---|---|---|---|---|---|---|
| Task 1 | 99.7163 | 91.8203 | 0 | 2.31678 | 0 | 1.513 | 0 | 2.83688 | 0 | 95.0827 |
| Task 2 | 0 | 61.8022 | 94.5642 | 55.191 | 0 | 1.56709 | 0 | 1.42018 | 0 | 61.6063 |
| Task 3 | 0.0533618 | 64.301 | 0 | 80.2561 | 98.1323 | 93.063 | 0.0533618 | 94.5037 | 0 | 56.4034 |
| Task 4 | 0 | 85.9517 | 0 | 83.3333 | 0 | 8.86203 | 99.144 | 74.7231 | 0.20141 | 85.7503 |
| Task 5 | 0 | 2.37015 | 0 | 0.706001 | 0 | 2.97529 | 0 | 0.605144 | 96.2179 | 51.0338 |
| All tasks | 21.1 | 61.63 | 19.31 | 43.49 | 18.39 | 20.43 | 19.7 | 33.56 | 19.12 | 70.41 |

Figure 3: Maximum accuracy ( 70%) on MNIST catastrophic forgetting task. Task 1 = digits 4 and 5, Task 2 = digits 6 and 7, Task 3 = digits 2 and 3, Task 4 = digits 0 and 1, Task 5 = digits 8 and 9.

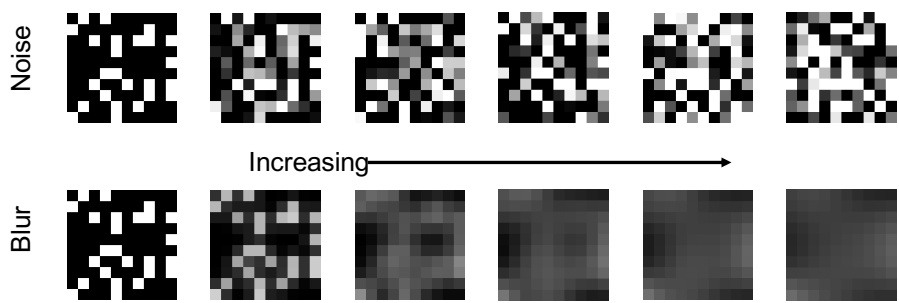

Figure 4: Types of images tested on for generalization for the Patches dataset. Top - Images with Gaussian noise added with increasing variannce (from 0 to 1.0 in steps of 0.2). Bottom - Gaussian blurred images with increasing sigma (from 0 to 2.5 in steps of 0.5).

3

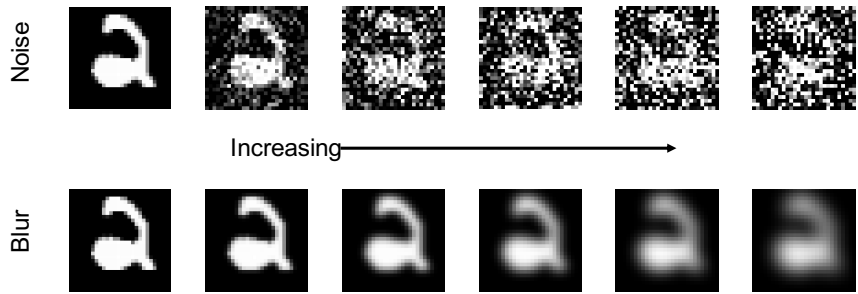

Figure 5: Types of images tested on for generalization for the MNIST dataset. Top - Images with Gaussian noise added with increasing variannce (0, 0.1, 0.3, 0.5, 0.7, 0.9, left to right). Bottom - Gaussian blurred images with increasing sigma (from 0 to 2.5 in steps of 0.5).


\end{document}